\documentclass[runningheads,a4paper]{llncs}

\usepackage{amssymb}
\usepackage{amsmath}
\DeclareMathOperator*{\argmaxC}{\arg\max}   
\usepackage{graphicx}
\usepackage{subfig}
\usepackage{float}
\usepackage{stfloats}

\usepackage{url}
\urldef{\mailsa}\path|{alfred.hofmann, ursula.barth, ingrid.haas, frank.holzwarth,|
\urldef{\mailsb}\path|anna.kramer, leonie.kunz, christine.reiss, nicole.sator,|
\urldef{\mailsc}\path|erika.siebert-cole, peter.strasser, lncs}@springer.com|    
\newcommand{\keywords}[1]{\par\addvspace\baselineskip
\noindent\keywordname\enspace\ignorespaces#1}

\begin{document}

\mainmatter  

\title{Quantifying the Uncertainty in Model Parameters using Gaussian Process-Based Markov Chain Monte Carlo: An Application to Cardiac Electrophysiological Models}

\titlerunning{Quantifying the Uncertainty in Model Parameters}

%
%

\author{Jwala Dhamala\textsuperscript{1} \and John L. Sapp\textsuperscript{2} \and Milan Horacek\textsuperscript{2} \and Linwei Wang\textsuperscript{1}%
}
%

\institute{\textsuperscript{1} Rochester Institute of Technology, Rochester, NY, 14623, USA\\
	\textsuperscript{2} Dalhousie University, Halifax, Canada
}

\toctitle{Lecture Notes in Computer Science}
\tocauthor{Authors' Instructions}
\maketitle

\begin{abstract}
Estimation of patient-specific model parameters is important for 
personalized modeling, 
although 
sparse and noisy clinical data 
can introduce significant uncertainty in the estimated parameter values. 
This importance source of uncertainty, 
if left unquantified, 
will lead to unknown variability in model outputs 
that hinder their reliable adoptions. 
Probabilistic estimation model parameters, 
however, 
remains an unresolved challenge 
because standard Markov Chain Monte Carlo sampling 
requires repeated model simulations that are computationally infeasible. 
A common solution 
is to replace the simulation model with a computationally-efficient surrogate for a faster sampling. 
However, by sampling from an approximation of the exact posterior probability density function (pdf) of the parameters, 
the efficiency is gained at the expense of sampling accuracy. 
In this paper, 
we address this issue by 
integrating surrogate modeling 
into Metropolis Hasting (MH) sampling of the exact posterior pdfs 
to improve its 
acceptance rate.   
It is done by first quickly constructing a Gaussian process (GP) surrogate of the exact posterior pdfs  
using deterministic optimization.   
This efficient surrogate is then used to modify commonly-used proposal distributions in MH sampling 
such that only proposals accepted by the surrogate 
will be tested by the exact posterior pdf for acceptance/rejection, 
reducing unnecessary model simulations at unlikely candidates. 
Synthetic and real-data experiments using the presented method 
show a significant gain in computational efficiency without compromising the accuracy. 
In addition, 
insights into the non-identifiability and heterogeneity 
of tissue properties can be gained from the obtained posterior distributions. 
\end{abstract}

\section{Introduction}
Patient-specific models are showing increasing promise 
in personalized medicine \cite{sermesant2012patient}. 
While advancement in medical imaging 
has made personalized geometrical models a reality, 
the challenge of obtaining patient-specific tissue properties 
in the form of model parameters 
remains unresolved. 
These model parameters often cannot be directly measured, 
but have to be inferred from clinical data 
that are sparse and noisy.

Most existing works on parameter estimation 
use deterministic methods to 
find an \emph{optimal} value of the model parameter 
so that model outputs best fit the measurement data~\cite{wong_strain-based_2012,sermesant2012patient,dhamala2016spatially}. 
However, significant uncertainty can exist in the estimated parameter values 
due to the uncertainty in available data. 
This important source of uncertainty will result in unknown variability in model outputs that, 
if left unquantified, 
will hinder their reliable adoptions. 
Additionally, over-parameterization and coupling between parameters 
may result in many parameter configurations that fit the data equally well. 
This issue of identifiability cannot be observed 
when only an \emph{optimal} solution is being sought.

A probabilistic estimation of model parameters can address the above challenges
 by obtaining the posterior probability density (pdf) of the parameters 
 conditioned on the data~\cite{schiavazzi2015uncertainty,konukoglu2011efficient,le2015bayesian}. 
 However, limited progress has been made in this direction 
 because the posterior pdf comprises of a complex simulation model 
 that are analytically intractable and computationally expensive. 
 While Markov Chain Monte Carlo (MCMC) methods are natural choices  
 for drawing samples from an analytically-intractable pdf, 
 they become prohibitive in this context 
 because the evaluation of each sample involves a model simulation that 
 could take hours or even days for a single run. 
 To address this critical challenge, 
 an effective approach is to construct an efficient surrogate 
 for the compute-intensive simulation model 
 using methods such as polynomial chaos~\cite{konukoglu2011efficient} and kriging~\cite{schiavazzi2015uncertainty}. 
 These surrogate models can then replace the original model in the posterior pdf 
 for substantially faster sampling~\cite{konukoglu2011efficient,schiavazzi2015uncertainty}. 
 However, this approach has two major limitations. 
 First, the sampling is carried out on an approximated rather than the exact posterior pdf.   
Thus the efficiency is gained at the expense of sampling accuracy. 
 Second, the surrogate is built to be accurate 
 in important regions of the simulation model \cite{konukoglu2011efficient,schiavazzi2015uncertainty}. 
Thus the approximation accuracy can be limited in important regions of the posterior pdf, such as those of high probability. 

We propose  
to overcome these issues by  
integrating surrogate modeling of the posterior pdf 
into the classic form of MCMC sampling 
to improve its acceptance rate. 
A similar idea was reported
in~\cite{le2015bayesian}, 
where 
a Gaussian process (GP) surrogate of the posterior pdf was constructed using hybrid Monte Carlo (HMC), 
the gradient of which was then used to find better proposals when sampling the exact posterior pdf. 
However, 
to construct the GP surrogate by random exploration of the sampling space, 
the number of model simulations needed 
increases exponentially with the number of unknown parameters. 
In addition, 
although the gradient of the GP surrogate 
allows a smarter exploration of the sampling space during HMC, 
the simulation model still needs to be probed at each proposed sample 
whereas a large portion of such heavy computation is spent at rejecting unwanted proposals. 
Important challenges remain 
in order to further reduce the number of model simulations needed in this type of approaches. 

In this paper, 
we address these challenges from both the end of surrogate modeling and MCMC sampling. 
First, 
rather than a random exploration, 
we quickly construct a GP surrogate of the posterior pdf by 
deterministic optimization 
favoring high accuracy in regions of high posterior probability. 
Second, 
we 
modify common proposal distributions in  Metropolis Hastings (MH) sampling 
by first testing the acceptance of each proposal on the far-cheaper-to-evaluate GP surrogate,  
quickly rejecting a large number of proposals and 
allowing only those accepted by the surrogate to be tested by the exact pdf. 
Compared to directly sampling from an approximated pdf \cite{konukoglu2011efficient,schiavazzi2015uncertainty}, 
the presented method ensures a high accuracy by generating the final samples from the exact distribution. 
Compared to directly sampling the exact pdf, 
the presented method reduces computation 
by 
avoiding expensive model simulations at unlikely proposals. 

We evaluate the presented method on estimating tissue excitability 
of a cardiac electrophysiological model 
using non-invasive  electrocardiogram (ECG) data. 
In synthetic experiments, 
we first evaluate the sampling accuracy and computational cost of the presented method 
against directly sampling the exact posterior pdf. 
Using the exact posterior pdf as a baseline, 
we then compare the accuracy of the presented method 
to an approach that, 
similar to existing work~\cite{schiavazzi2015uncertainty}, 
samples only the surrogate posterior pdf. 
Finally, in both synthetic and real-data studies, 
we analyze the uncertainty, identifiability, and heterogeneity 
of tissue excitability using its posterior pdfs personalized from ECG data.

\section{Cardiac Electrophysiological System}
\label{sec:models}

\noindent{\textbf{Whole-heart Electrophysiology Model:}}
Simplified electrophysiological models are popular candidates 
for parameter estimation in personalized modeling~\cite{sermesant2012patient}. 
They 
can reproduce the general shape of action potential with 
a small number of parameters and reasonable computation. 
As a preliminary demonstration of the presented method, 
we consider the two-variable~\emph{Aliev-Panfilov} (AP) model~\cite{aliev_simple_1996}:
\begin{eqnarray}
		{\partial u}/{\partial t} &=& {\partial}/{\partial x_i}d_{ij}{\partial u}/{\partial x_j} - ku(u-a)(u-1) -uv, \nonumber\\
		{\partial v}/{\partial t} &=& 	\varepsilon(u,v)(-v -ku(u-a-1)),
	\label{eq:AP}
\end{eqnarray}
where $\mathit{u}$ is the action potential and 
$\mathit{v}$ is the recovery current. 
Parameter $\mathit{d_{ij}}$ is the conductivity, 
parameter $\mathit{\varepsilon}$ controls the coupling between the recovery current and action potential, 
$\mathit{k}$ controls the repolarization, 
and $\mathit{a}$ controls the excitability of the myocyte. 
As a proof of concept, 
in this study we consider quantifying the uncertainty of parameter $a$ 
because  it is closely associated with the ischemic severity of the myocardial tissue 
and model output $u$ is sensitive to its value. 

The meshfree method as described in~\cite{wang2010physiological} 
is used to discretize and solve the AP model on the 3D myocardium. 
The direct estimation of parameter $a$ 
at the resolution of the cardiac mesh 
is impossible due to non-identifiability and heavy computation. 
Instead, we consider the estimation of parameter $a$ at a reduced dimension. 
Any appropriate dimension reduction method can be used. 
Here, we use a recently reported method 
that automatically groups spatial nodes of the cardiac mesh into $5-17$ regions that are at various resolutions and that are most homogeneous in tissue properties~\cite{dhamala2016spatially,dhamala2017spatially}.\\

\noindent{\textbf{Measurement Model:}} 
Spatio-temporal cardiac action potential 
produces time-varying ECG signals on the body surface. 
This relationship can be described by the quasi-static approximation of the electromagnetic theory~\cite{plonsey1969bioelectric}. Solving the governing equations
on a discrete mesh of heart and torso~\cite{wang2010physiological}, 
a linear model between ECG data $\mathbf{Y}$ and action potential $\mathbf{U}$ 
can be obtained as: $\mathbf{Y} = \textbf{H}\textbf{U}(\pmb{\theta}).$ 
Here, $\mathbf{H}$ is the transfer matrix unique to each heart and torso geometry, 
and $\pmb{\theta}$ is the vector of spatially-varying local parameters $a$ 
at a reduced dimension.

\section{Probabilistic Parameter Estimation}
The relationship between parameter $\pmb{\theta}$ 
and ECG data $\textbf{Y}$ can be expressed as:
\begin{equation}
\mathbf{Y} = F(\pmb{\theta}) + \pmb{{\epsilon}} 
\label{eq:relationship}
\end{equation}
where $F$ 
consists of the whole-heart electrophysiological model 
and the measurement model as described in section~\ref{sec:models}. 
$\pmb{{\epsilon}}$ 
accounts for discrepancy between model outputs and measurement data. 
Using Bayes' theorem, 
the unnormalized posterior density of the model parameter $\pmb{\theta}$ can be obtained as:
\begin{equation}
\pi(\pmb{\theta}|\mathbf{Y}) \propto \pi(\mathbf{Y}|\pmb{\theta})\pi(\pmb{\theta})
\label{eq:pos}
\end{equation}
Assuming noise $\pmb{\epsilon}$ to follow a zero mean Gaussian distribution 
with a diagonal covariance $\mathbf{\Sigma}_e=\sigma_e^2\mathbf{I}$, 
the likelihood $\pi(\mathbf{Y}|\pmb{\theta})$ can be formulated as:
\begin{equation}
\pi(\mathbf{Y}|\pmb{\theta}) \propto \mathrm{exp}(-1/2\sigma_e^2 ||\mathbf{Y}-F(\pmb{\theta})||^2) 
\label{eq:lik}
\end{equation}
The prior distribution $\pi(\pmb{\theta})$ quantifies prior knowledge over the parameter. 
Here we use a uniform distribution in bounded space $[0,0.52]$ 
to include the minimal physiological knowledge 
about the range of values for this parameter:  
$a\sim0.15$ represents normal excitability, 
while increased value represents increased loss of excitability 
until $a\sim0.5$ represents necrotic tissue.

MCMC sampling of the posterior pdf in equation~(\ref{eq:pos}) 
is infeasible because the evaluation of each sample requires an expensive model simulation. 
Below we describe the presented method 
that accelerates MCMC sampling of~(\ref{eq:pos}) 
via the use of an efficient GP surrogate in the modification of proposal distributions. \\

\noindent{\textbf{GP Surrogate of Posterior Functions:}} 
GP is a popular method to model a function 
that lacks an explicit form or is difficult to evaluate. 
such as a deep learning model, 
an experiment to be designed, 
and a multiscale simulation model~\cite{rasmussen2006gaussian}. 
Here, we use it to approximate the exact posterior pdf 
in equation~(\ref{eq:pos}). 

To initialize, we take a GP with a zero mean function 
and an anisotropic \textquotedblleft M\'{a}tern 5/2" co-variance function~\cite{rasmussen2006gaussian}:  
\begin{align}
\kappa(\pmb{\theta}_1,\pmb{\theta}_2) & = \alpha^2 \big\{1+\sqrt{5d^2(\pmb{\theta_1},\pmb{\theta_2})}+5/3d^2(\pmb{\theta}_1,\pmb{\theta}_2) \textrm{exp} (-\sqrt{5d^2(\pmb{\theta}_1,\pmb{\theta}_2)}\big\} 
\end{align}
where $d^2(\pmb{\theta}_1,\pmb{\theta}_2) = (\pmb{\theta}_1-\pmb{\theta}_2)^\intercal\mathbf{\Lambda}(\pmb{\theta}_1-\pmb{\theta}_2)$, 
the diagonal of $\mathbf{\Lambda}$ are length scales, 
and $\alpha^2$ is the co-variance amplitude. 
This kernel 
relaxes the assumption on the smoothness of the posterior pdf 
compared to commonly-used squared exponential kernel. 
The GP is then learnt by 
an iteration of the following two steps: 

\emph{1. Finding optimal points to build the GP:} 
Sample points used to build the GP 
are placed so as to: 
1) globally approximate the posterior pdf, 
and 2) concentrate more in regions of high posterior probability. 
For the former, 
points are chosen where the predictive uncertainty $\sigma(\pmb{\theta}$) of the current GP is high 
(to facilitate exploration of uncertain space). 
For the latter, 
points are chosen where the predictive mean $\mu(\pmb{\theta}$) of the current GP is high 
(to exploit current knowledge about the space of high posterior probability). 
This is done by finding the point that 
maximizes the upper confidence bound of the GP~\cite{rasmussen2006gaussian}: 
\begin{align}
\hat{\pmb{\theta}} & = \argmaxC_{\pmb{\theta}}\big\{\mu(\pmb{\theta}) + \beta^{1/2} \sigma(\pmb{\theta})\big\}
\label{eq:ucb}
\end{align}
where $\mu(\pmb{\theta}$) and $\sigma(\pmb{\theta}$) 
are evaluated by the Sherman-Morrison-Woodbury formula~\cite{rasmussen2006gaussian}. 
The parameter $\beta$
balances between exploitation and exploration of the sample space~\cite{rasmussen2006gaussian}. 
Equation~(\ref{eq:ucb}) is optimized using 
Bound Optimization BY Quadratic Approximation (BOBYQA)~\cite{powell2008developments}. 

\emph{2. Updating the GP  surrogate:} Once a new point is obtained, 
the exact posterior pdf~(\ref{eq:pos}) is evaluated at this point 
and 
the GP is updated. 
After every a few updates of the GP, 
we optimize the hyperparameters 
(length scales $\mathbf{\Lambda}$ and covariance amplitude $\alpha$)  
by maximizing the marginal likelihood. 

These two steps iterate until new points collected change little. 
In this way, 
we quickly obtain 
a surrogate $\pi^*(\pmb{\theta}|\mathbf{Y})$ of the exact posterior pdf~(\ref{eq:pos}) 
that is much cheaper to evaluate and is most accurate in regions of high posterior probability.\\

\noindent{\textbf{MCMC Acceleration using GP Approximation:}} 
Metropolis Hasting (MH) is 
commonly used 
for generating a Markov chain of samples from a stationary distribution~\cite{adrieu2003introduction}.  
Supposing that the $n^{th}$ sample in the Markov chain is $\pmb{\theta}_n$, 
MH in its native form 
first draws a random candidate from a proposal distribution $q(\pmb{\theta}|\pmb{\theta}_n)$, 
and then accepts the candidate with an acceptance probability given as:
\begin{equation}
\rho(\pmb{\theta}_n,\pmb{\theta}) = \textrm{min}\Big(1,\frac{q(\pmb{\theta}_n|\pmb{\theta})\pi(\pmb{\theta}|\mathbf{Y})}{q(\pmb{\theta}|\pmb{\theta}_n)\pi(\pmb{\theta}_n|\mathbf{Y})}\Big)
\label{eq:mh}
\end{equation}
which means that the expensive posterior pdf~(\ref{eq:pos}) has to be probed 
--- namely, the simulation model to be run --- 
at every proposed sample. 
Because obtaining a proposal $q(\pmb{\theta}|\pmb{\theta}_n)$ similar to the exact posterior pdf is notoriously difficult~\cite{adrieu2003introduction,gilks1995markov}, 
the acceptance rate is often low which leads to 
an infeasible amount of computation that will mainly be spent at rejecting unwanted proposals. 

To improve the acceptance rate of MH, 
we use 
the much-faster-to-evaluate GP surrogate to modify commonly-used proposal distributions, 
such as a Gaussian distribution centered at $\pmb{\theta}$ 
with covariance $\mathbf{\Sigma}_p = \sigma^2_p\mathbf{I}$. 
A candidate sample $\pmb{\theta}$ is first drawn 
from the proposal distribution as usual. 
Instead of directly testing the acceptance of this candidate 
using equation~(\ref{eq:mh}), 
we first test the acceptance of this candidate against the surrogate GP $\pi^*(\pmb{\theta}|\mathbf{Y})$ using a probability $\rho_A$: 
\begin{equation}
\label{eq:gp_mh}
\rho_A(\pmb{\theta}_n,\pmb{\theta}) = \textrm{min}\Big(1,\frac{q(\pmb{\theta}_n|\pmb{\theta})\pi^*(\pmb{\theta}|\mathbf{Y})}{q(\pmb{\theta}|\pmb{\theta}_n)\pi^*(\pmb{\theta}_n|\mathbf{Y})}\Big) 
\end{equation}
which is computationally much cheaper to evaluate compared to~(\ref{eq:mh}).  
Intuitively, 
this additional stage of acceptance or rejection 
modifies the 
proposal distribution by filtering out 
candidates that have a high probability of being rejected by the exact pdf. 
Only the candidates accepted by the GP surrogate 
are then evaluated against the exact posterior pdf 
for acceptance with a probability $\rho_E$: 
\begin{equation}
\rho_E(\pmb{\theta}_n,\pmb{\theta}) = \textrm{min}\Big(1,\frac{\pi(\pmb{\theta}|\mathbf{Y})\pi^*(\pmb{\theta}_n|\mathbf{Y})}{\pi(\pmb{\theta}_n|\mathbf{Y})\pi^*(\pmb{\theta}|\mathbf{Y})}\Big) 
\end{equation}
In this way, 
the acceptance rate is improved and 
unnecessary model simulations are avoided 
at proposals that would have been rejected with high probability. 
This is achieved without sacrificing sampling accuracy 
because the final Markov chain is generated via acceptance 
by the exact posterior pdf. 
Due to space limit, 
refer to~\cite{christen2012markov,efendiev2006preconditioning} for 
discussions on the ergodicity 
and convergence of the Markov chain to the exact pdf when modifying the proposal using an approximation. 

\section{Experiments}
\noindent{\textbf{Synthetic Experiments:}}
On two image-derived human heart-torso models, 
we include six cases of infarcts of different sizes 
and locations of the LV. 
Note that 
a relatively small number of experiments is considered 
because it is time consuming 
to obtain samples on the exact posterior pdf as a baseline. 
For each synthetic case, 
parameter $a$ in the AP model 
is set to be $0.15$ for normal tissue and $0.5$ for infarct tissue. 
120-lead ECG is simulated and corrupted with 20dB Gaussian noise as measurement data. 
After dimensionality reduction~\cite{dhamala2016spatially},  
the number of parameters to be estimated in each case 
ranges from 9 to 12. 

In all MH sampling, 
we use a four parallel chains with same Gaussian proposal distribution but four different initial points.
Because the GP surrogate is efficient to sample, 
we use it to tune 
the variance of the proposal distribution 
and the starting point for each MCMC chain.
The former is tuned 
to attain an acceptance rate of $0.3-0.4$. 
For the latter, 
a rapid sampling of the GP surrogate is first conducted  
using slice sampling for 20,000 samples; 
assuming that these samples come from a mixture of four Gaussian distributions, 
the mean of each is then used to start each of the four parallel chains. 
After discarding initial burn-in samples 
and picking alternate samples to avoid auto-correlation in each chain, 
the samples from four chains are combined. 
The convergence of all MCMC chains are tested using 
trace plots, Geweke statistics, 
and Gelman-Rubin statistics~\cite{adrieu2003introduction,gilks1995markov}. 
\begin{figure}[!t]
	\centering
 		\includegraphics[width=\textwidth]{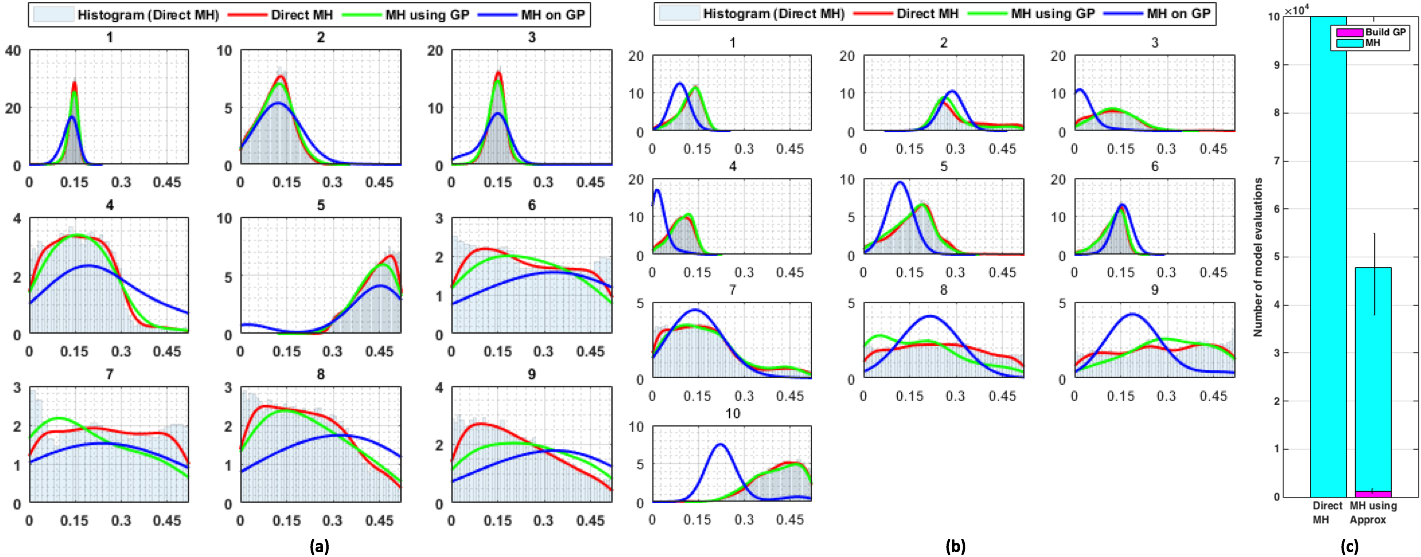}
    \caption{
    (a)-(b): Examples of exact posterior pdfs (red) \emph{vs}.\ those obtained by the presented method (green) and sampling the GP surrogate only (blue). 
    (c): Comparison of efficiency in terms of the number of model simulations between the presented method (right bar) and sampling the exact posterior pdf (left bar).} 
    \label{fig:comparision}
\end{figure}

\emph{Validation against exact posterior pdf:} 
Fig.~\ref{fig:comparision}(a)-(b) presents examples of posterior pdfs 
obtained from two experiments 
where the dimension of unknown parameters are 9 and 10, respectively. 
As shown, 
the presented sampling strategy (green curve) closely reproduces true posterior pdf (red curve). 
In the mean time, 
it reduces the computational cost by an average of $53.56\%$ 
(Fig.~\ref{fig:comparision}(c)) 
despite the overhead of constructing the GP surrogate which, as highlighted in the purple bar in Fig.~\ref{fig:comparision}(c), is negligible compared to the computation required for sampling.

\begin{table}[!t]
\centering
\caption{
Absolute errors in mean, mode, and standard deviation  
against the exact posterior pdf: 
the presented method \emph{vs}.\ directly sampling the GP surrogate.}
\label{table}
\begin{tabular}{@{}llll@{}}
\hline                                                             & \multicolumn{1}{c}{\textbf{Mean}} & \multicolumn{1}{c}{\textbf{Mode}} & \multicolumn{1}{c}{\textbf{Standard deviation}} \\ \hline
\begin{tabular}[c]{@{}l@{}}\textbf{Presented method}\end{tabular}   & 0.0154$\pm$0.0186                 & 0.0510$\pm$0.0711                 & 0.0059$\pm$0.0074                \\
\begin{tabular}[c]{@{}l@{}}\textbf{Sampling surrogate}\end{tabular} & 0.0549$\pm$0.0532                 & 0.0972$\pm$0.1111                 & 0.0309$\pm$0.0306                \\ \hline
\end{tabular}
\end{table}

\emph{Comparison with directly sampling the GP surrogate:} 
Directly sampling the GP posterior pdf in replacement of the exact pdf, 
as commonly done in existing methods~\cite{schiavazzi2015uncertainty}, 
requires significantly less computation because no model simulation is needed. 
However, the sampling accuracy is limited. 
This is especially evident in Fig.~\ref{fig:comparision}(b) 
where sampling the GP surrogate (blue curve) produces a distribution 
that is different from the exact pdf not only in general shape but 
also in locations of the mode. 
Using the  mean, mode, and standard deviation (std) of the exact pdf as a baseline, 
Table~\ref{table} shows that sampling errors of the presented method 
are significantly lower than those from sampling the surrogate 
(paired \emph{t}-test on 60 estimated parameters, $p <0.0012$).

\begin{figure}[!t]
	\centering
 		\includegraphics[width=0.8\textwidth]{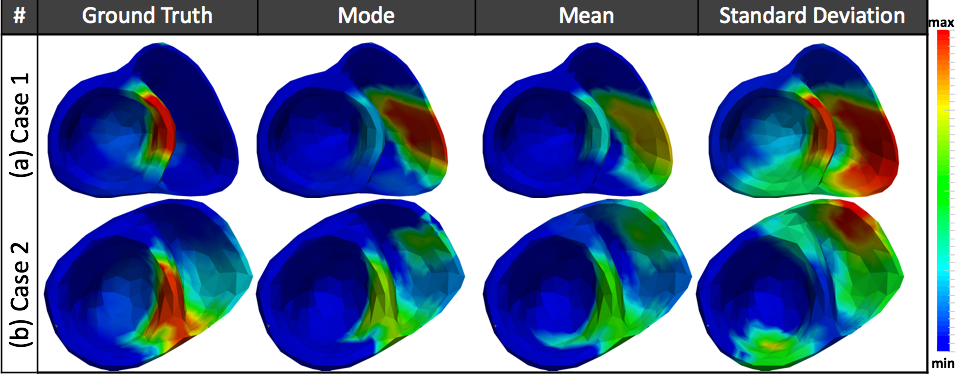}
    \caption{Synthetic data: mean, mode and std of estimated posterior pdfs.} 
    \label{fig:synResults}
\end{figure}

\begin{figure}[!t]
	\centering
 		\includegraphics[width=0.9\textwidth]{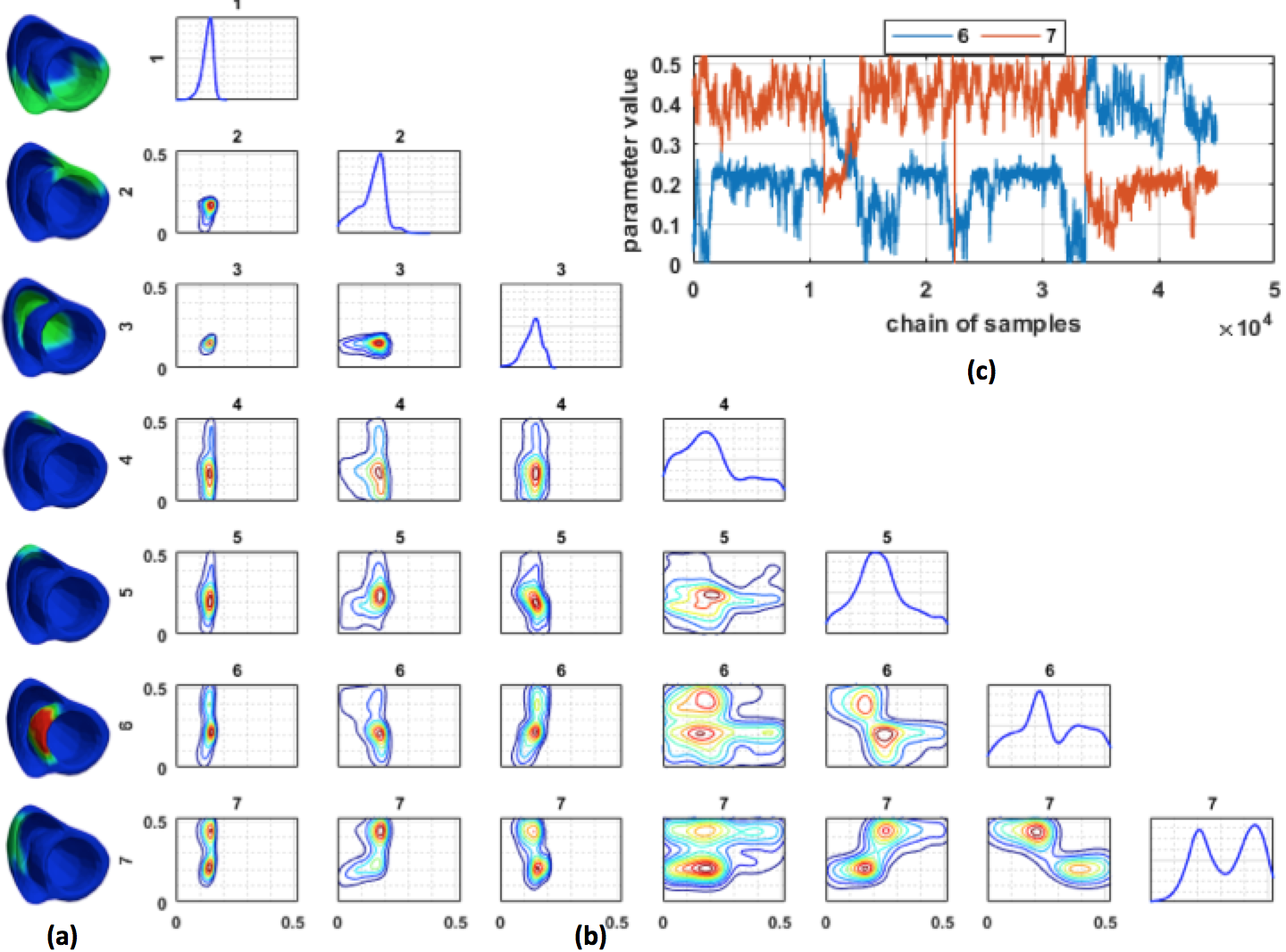}
    \caption{(a) Regions of heart to be parameterized 
    (red: infarct, green: non-infarct/mixed).  
    (b) Uni-variate and bi-variate marginal pdf plots. 
    (c) Trace plot for parameters of regions 6 and 7 showing switching behavior.} 
    \label{fig:sep2pairs}
\end{figure}

\emph{Analysis of uncertainty $\&$ identifiability:} 
Fig.~\ref{fig:synResults} shows maps of summary statistics 
obtained by the presented method in two cases with septal infarcts. 
In case 1, there are 7 regions of the heart to be parameterized. 
As shown, 
a strong false positive at the RV lateral wall is present at both the 
posterior mode and mean of the estimated parameters. 
The std map  
indicates that this false positive is associated with a high uncertainty. 
For a closer look at the reason for such uncertainty, 
Fig.~\ref{fig:sep2pairs} shows posterior pdfs of the estimated parameters 
(the region on which each parameter is estimated is shown in the first column, 
where green indicates a healthy region and red an infarcted region). 
As shown, 
the first three regions correspond to healthy regions 
and their parameters are estimated with a prominent single mode and low uncertainty. 
Regions 4 and 5 correspond to two small healthy regions close to the infarct. 
Their parameters are estimated accurately but with higher uncertainty. 
The last two regions correspond to the infarcted region and the RV wall. 
Their parameters show  a coupling that results in a bimodal distribution. 
This also exhibits as a switching behavior 
in the trace plot of these two parameters (Fig.~\ref{fig:sep2pairs}(c)): 
if the parameter of region 6 is estimated towards a healthier state, 
the parameter of  region 7 would tend towards an infarcted state. 
Such switching property in Markov chains is associated with non-identifiability~\cite{siekmann2012mcmc}, 
\emph{i.e.}, either combination of the two parameter values could fit the measurement data similarly well. 
Knowledge about this non-identifiability 
is valuable for placing proper trust 
in the obtained posterior point estimates.  

\begin{figure}[!t]
	\centering
 		\includegraphics[width=0.8\textwidth]{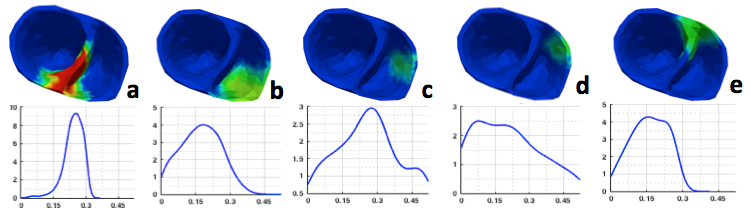}
    \caption{Uni-variate marginal probability density plots 
    and the corresponding regions of heart to be parameterized 
    for case 2 from Fig.~\ref{fig:synResults}.} 
    \label{fig:sep3density}
\end{figure}
In case 2, there are 11 regions to be parameterized, 
five of which 
along with their estimated parameter pdfs are shown in Fig.~\ref{fig:sep3density}. 
As shown, 
the parameter of the region that contains the true infarct (a)  
is correctly estimated with a narrow uni-modal distribution. 
In comparison, 
several RV regions adjacent to the septal infarct (b-e) 
have difficulty converging 
which we suspect could be again caused by non-identifiability. 
As a result, 
we obtain a solution where the true septal infarct is estimated with high confidence, 
whereas the false positives are associated with a higher uncertainty  
as summarized in Fig.~\ref{fig:synResults}. 
Namely, uncertainty analysis 
helps differentiate false positives from true positives in this case.\\


\begin{figure}[!t]
	\centering
 		\includegraphics[width=0.8\textwidth]{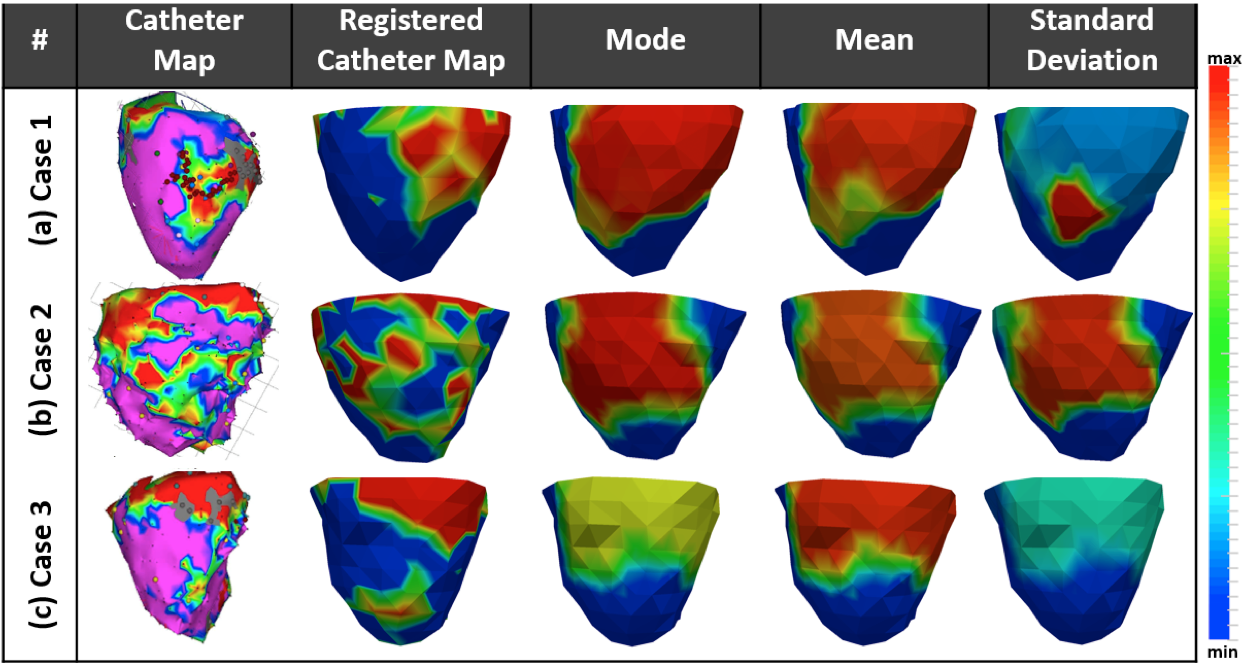}
   \caption{Real-data experiments: mean, mode and std of posterior pdfs.} 
    \label{fig:real_fig}
\end{figure}

\noindent{\textbf{Real-data Experiments:}} 
We conduct real-data studies on three patients 
who underwent catheter ablation of ventricular tachycardia due to prior infraction \cite{sapp2012inverse}. 
Patient-specific heart-torso geometrical models are obtained from axial CT images. 
The uncertainty of tissue excitability in the AP model (1) is estimated from 120-lead ECG data. For evaluation of the results, 
bipolar voltage data from \emph{\textit{in-vivo}} catheter mapping are used. 
However, it should be noted that voltage maps are not a direct measure of tissue excitability 
and thus should be interpreted as a reference but not the validation data. 
Fig.~\ref{fig:real_fig} shows the catheter data along with the estimation results, 
where the first column shows the original voltage maps (red: dense scar $\le$0.5mV; purple: healthy tissue $>1.5$mV; green: scar border $0.5-1.5$mV) 
and the second column shows the same voltage data 
registered to CT-derived cardiac meshes. 
As shown, 
compared to synthetic data, 
a real infarct is often  
distributed with higher heterogeneity. 
The resolution to which such heterogeneity can be captured 
is largely limited by the method of dimensionality reduction. 
Below we show how uncertainties of the lower-resolution estimation   
are associated with the heterogeneity of the underlying tissue. 

\begin{figure*}[!t]
	\centering
    \subfloat{\includegraphics[width=0.8\textwidth]{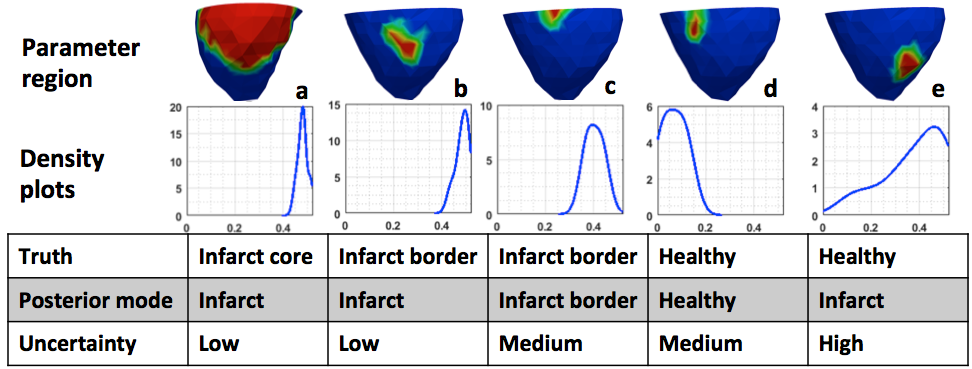}}%
    \caption{Real-data experiments: marginal probability density plots 
    and the corresponding regions of heart being parameterized in case 1.}
    \label{fig:dens_dc}
\end{figure*}

\emph{Case 1: }
The voltage data for 
case 1 (Fig.~\ref{fig:real_fig}(a)) shows a dense infarct at inferolateral LV 
with a heterogeneous region extending to lateral LV. 
Dimensionality reduction  
generates 12 regions of the heart to be parameterized, 
five of which are listed in Fig.~\ref{fig:dens_dc} 
along with the estimated posterior marginal pdfs 
for their parameters. 
As shown, 
the parameter for the region of infarct core (a) 
is correctly estimated with low uncertainty. 
For several regions around 
the heterogeneous infarct border (b-d), 
uncertainties of the estimation become higher. 
A particularly high uncertainty is obtained 
at a small healthy region by the scar border (e), 
where the parameter is incorrectly estimated. 
This produces an estimation 
with correct posterior mode/mean and low uncertainty 
at the infarct core, 
increased uncertainty at the heterogeneous infarct border, 
and high uncertainty at  a region of 
false-positive near the infarct border 
as summarized in Fig.~\ref{fig:real_fig}(a). 
\begin{figure*}[!t]
	\centering
		\subfloat{\includegraphics[width=0.8\textwidth]{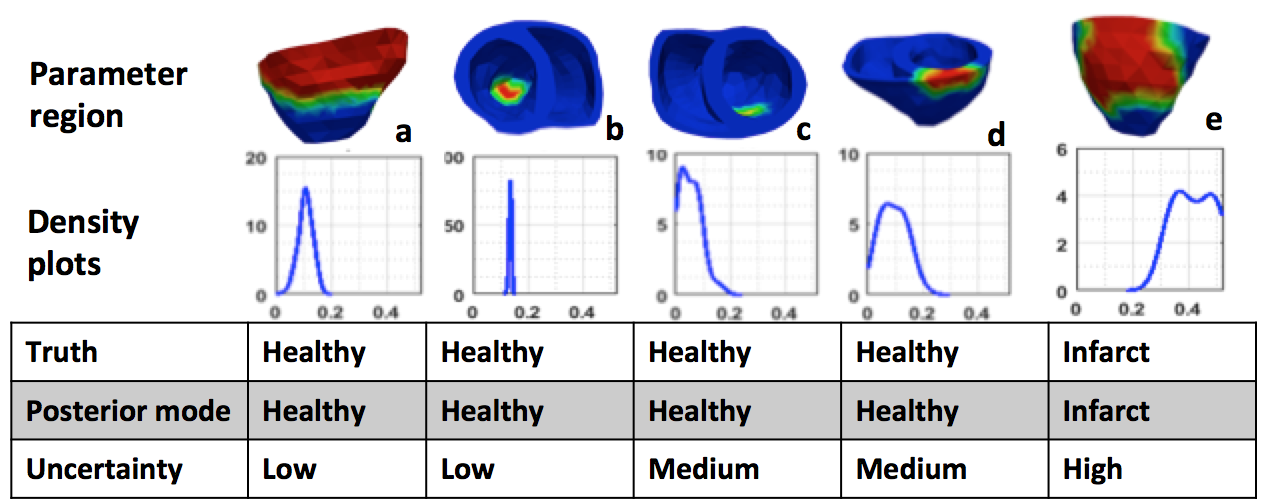}}%
    \caption{Real-data experiments: marginal probability density plots 
    and the corresponding regions of heart being parameterized in case 2.}
    \label{fig:dens_ec}
\end{figure*}

\emph{Case 2: } 
The voltage data for 
case 2 (Fig.~\ref{fig:real_fig}(b)) shows a 
massive yet quite heterogeneous infarct at lateral  LV. 
Dimensionality reduction  
generates 8 regions of the heart to be parameterized, 
five of which are listed in Fig.~\ref{fig:dens_ec} 
along with the estimated posterior marginal pdfs  
for their parameters. 
As shown, 
for healthy regions remote from the infarct (a-b), 
their parameters are correctly estimated with high confidence. 
For healthy regions close to the infarct (c-d), 
their parameters are correctly estimated but with lower confidence. 
For the region that corresponds to the infarct (e), 
its abnormal parameter is correctly captured but 
with a high uncertainty -- 
likely reflecting the heterogeneous nature of tissue properties 
in this region. 
As summarized in  Fig.~\ref{fig:real_fig}(b), 
while the estimation 
correctly reveals the region of infarct as in case 1, 
it is also associated with a higher uncertainty 
compared to the less heterogeneous infarct in case 1. 
\begin{figure}[!t]
	\centering
 		\includegraphics[width=0.8\textwidth]{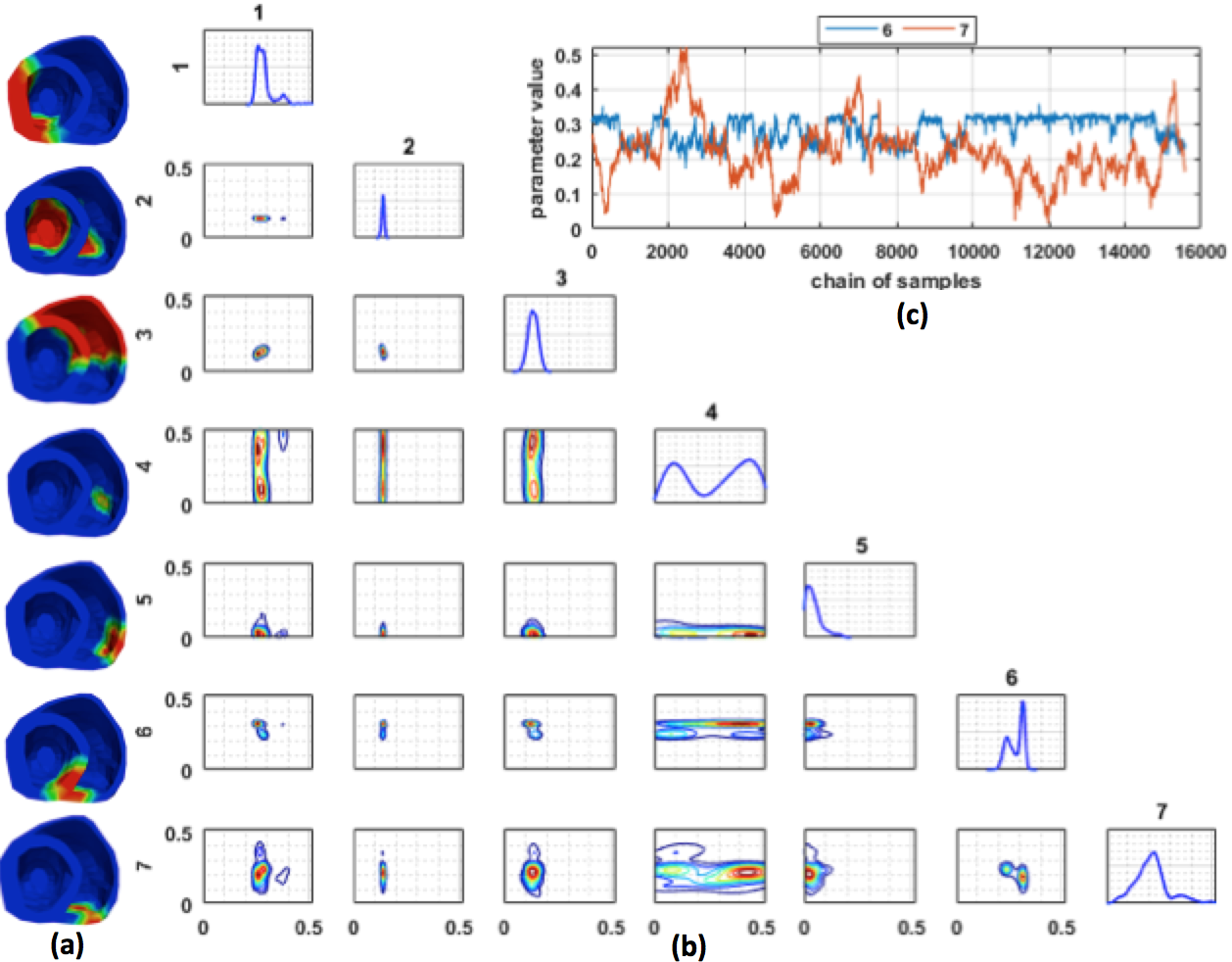}
    \caption{Real-data experiments for case 3. (a) Regions of  heart to be parameterized. 
    (b) Uni-variate and bi-variate marginal pdf plots. 
    (c) Trace plot for parameters of regions 6 and 7.} 
    \label{fig:aw_den}
\end{figure}

\emph{Case 3:}
The catheter data for 
case 3 (Fig.~\ref{fig:real_fig}(c)) 
shows low voltage at 
lateral LV and RV, 
although it was not clear whether
the low voltage on lateral RV 
was due to an infarct or fat layer. 
As shown in Fig.~\ref{fig:aw_den}, 
there are 7 regions of the heart to be parameterized. 
The abnormal parameter in lateral LV (region 1)  
is estimated with a narrow uni-modal distribution. 
In contrast, 
the marginal distribution for the parameter in lateral RV (region 4) 
shows a bimodal distribution with one mode in healthy range 
and the other in infarcted range. 
Markov chains for the parameters of two nearby regions at RV (regions 6-7) 
also show a switching behavior (Fig.~\ref{fig:aw_den}(c)). 
This produces an estimate of abnormal tissue property with high confidence at lateral LV (Fig.~\ref{fig:real_fig}(c)), 
and less confidence at lateral RV.

\section{Conclusion}
This paper presents 
a novel approach to 
efficiently yet accurately sample 
the distribution of parameters in complex simulation models. 
This is achieved by using GP-based surrogate modeling 
to improve the proposal distribution. 
A more accurate GP surrogate of the posterior pdf 
is more expensive to build but 
more effective in improving the acceptance rate of MH sampling, 
 while a less accurate GP surrogate is faster to build 
 but less effective in accelerating MH sampling. 
How to maintain this balance is to be investigated in future works.\\

{\textbf{Acknowledgments. }This work is supported by the National Science Foundation under CAREER Award ACI-1350374 and the National Institute of Heart, Lung, and Blood of the National Institutes of Health under Award R21Hl125998.}
\bibliographystyle{splncs03}
\bibliography{main}
\end{document}